\documentclass{article}
\usepackage{spconf,amsmath,graphicx,hyperref}
\usepackage{amsmath,amssymb}
\usepackage{color}
\usepackage{booktabs}
\usepackage{indentfirst} %
\usepackage{diagbox}
\usepackage{caption}

\usepackage{colortbl}
\usepackage{xcolor}
\definecolor{RowColor}{rgb}{0.93, 0.93, 1}

\usepackage{balance}

\title{Collaborative Compression for Large-Scale MoE Deployment on Edge}

\name{
  \parbox{\textwidth}{\centering
  Yixiao Chen$^{1}$, Yanyue Xie$^{1}$, Ruining Yang$^{1}$,Wei Jiang$^{2}$, Wei Wang$^{2}$,\\
 Yong He$^{2}$, Yue Chen$^{2}$, Pu Zhao$^{1*}$, Yanzhi Wang$^{1}$\thanks{*Corresponding authors}}
}
\address{
  \begin{tabular}{c c}
    $^{1}$Northeastern University
    $^{2}$Futurewei Technologies, Inc. \\
    \{chen.yixia, p.zhao, yanz.wang\}@northeastern.edu
  \end{tabular}
}

\begin{document}
\ninept
\maketitle

\begin{abstract}
The Mixture of Experts (MoE) architecture is an important method for scaling Large Language Models (LLMs). It increases model capacity while keeping computation cost low. 
However, the ultra-large MoE models still have hundreds of billions of parameters, requiring massive memory/storage and leading to difficulties for deployment on resource-constrained edge platforms.
Pruning or quantization alone can hardly address the issue, because of the super-aggressive compression ratio with significantly degraded accuracy and output quality. 
To facilitate the deployment of ultra-large MoEs on edge platforms, we propose a collaborative compression framework by combining 
expert pruning, mixed-precision quantization, and activation optimization.
It can effectively reduce the storage footprint of the ultra-large MoE DeepSeek-V3 from 1.3TB to 103GB, while preserving high output quality with better accuracy than traditional uniform low-bit quantization methods. 
To the best of our knowledge, we are the first to deploy a compressed model from the ultra-large DeepSeek-V3 on the platform with a strict 128GB total memory limit. 
Our comprehensive experiments on multiple benchmarks under various memory constraints demonstrate the effectiveness of our method with smaller model sizes and higher accuracy than uniform low-bit quantization methods. All our models are available at 
\url{https://huggingface.co/bobchenyx/DeepSeek-V3-0324-MLA-GGUF}
.

\end{abstract}

\section{Introduction}

The Mixture of Experts (MoE) architecture~\cite{jacobs1991adaptive,shazeer2017outrageously} has proven to be a pivotal technique for scaling Large Language Models (LLMs). By activating only a subset of experts for each input, MoE models can significantly increase capacity while keeping the computational cost relatively low. Recent works~\cite{lepikhin2020gshard,fedus2022switch,jiang2024mixtral,liu2024deepseek} have demonstrated the advantage of sparse activation in balancing parameter scale expansion and inference efficiency. However, despite the reduced computational complexity, the latest generation of ultra-large MoE models~\cite{liu2024deepseek,team2025kimi,qwen3technicalreport} still contain hundreds of billions of parameters  with terabyte-level memory requirements. 
Deploying such large models is difficult and requires substantial GPU and CPU memory on resource-constrained edge platforms. 

Various optimization strategies for MoE models are explored to lower memory usage during inference, including expert routing optimization~\cite{fedus2022switch,lepikhin2020gshard,rajbhandari2022deepspeed,he2022fastermoe,hwang2023tutel}, low-bit quantization~\cite{frantar2022gptq,lin2024awq,lee2024owq,ma2024affinequant,llama_cpp} and activation compression~\cite{xiao2023smoothquant,wei2022qdrop,dao2022flashattention,dao2023flashattention,ren2021zero,shen2024lazydit,zhan2024fast,chen2016training}.   However,  one single strategy can hardly satisfy the memory constraints for inference on edge platforms. Specifically, quantization is commonly adopted to save memory. However, even the extremely low-bit quantization (such as the uniform low-bit quantization down to IQ1\_S with an average bit width of 1.56 in llama.cpp~\cite{llama_cpp}) can only reduce the model storage to around 125GB, which still makes inference difficult on edge platforms with a total of 128GB memory across GPU and CPU. Furthermore, the generation quality suffers from the extremely low-bit quantization with unusable garbled text output. Meanwhile, pruning needs to remove substantial weights to reduce memory cost from terabyte-level to 128GB, resulting in severe performance degradation. 

This suggests that a single strategy is insufficient to ensure feasible inference under strict hardware constraints for MoEs. It is necessary to incorporate multiple compression strategies for jointly reducing both the weight storage and memory consumption of intermediate activations. However, there is a lack of systematic joint optimization research to combine multiple strategies for addressing  both weight and activation memory bottlenecks, thus facilitating  edge deployment of ultra-large MoEs  with hundreds of billions of weights.

To address these challenges, we propose a collaborative compression framework for large-scale MoE models that integrates expert pruning, mixed-precision quantization, and activation optimization. Our framework significantly reduces both weight and activation memory while preserving core model capabilities. 
To the best of our knowledge, we are the first to enable successful deployment of ultra large scale MoEs such as  DeepSeek-V3  under a 128GB memory limit. 
Furthermore, it outperforms uniform low-bit quantization methods by achieving higher accuracy across various benchmarks with lower memory cost.

The contributions of this work are summarized as follows:
\vspace{-0.5em}
\begin{itemize}
    \item We develop a compression framework for large-scale MoE models that supports practical deployment of DeepSeek-V3-scale models within a 128GB memory budget while maintaining inference usability and stable performance.
    \item We design and validate a cooperative optimization strategy that combines expert pruning with mixed-precision quantization, achieving substantial weight storage reduction while avoiding the severe accuracy degradation.%
    \item We introduce activation optimization techniques to lower peak memory usage during inference, further improving deployability in resource-constrained environments.
    \item Our experiments demonstrate that our method can effectively reduce model sizes with higher accuracy across various benchmarks, compared with existing low-bit quantizations.
\end{itemize}

\section{Related Work}

\begin{figure*}[t] 
    \centering
    \includegraphics[width=\linewidth]{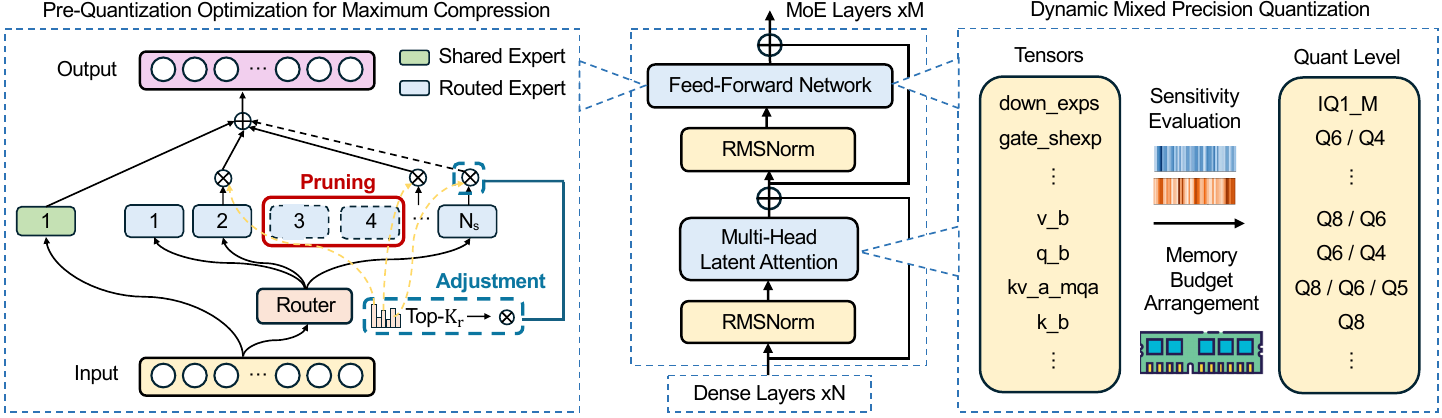}
    \caption{Illustration of our proposed MoE compression framework. 
    The middle part shows a block-level schematic of DeepSeek MoE. 
    The left part highlights our Performance-Aware Expert Reduction and Pruning-Aware Activation Adjustment for FFN layers, 
    while the right part illustrates the mixed-precision quantization process applied to the remaining model.
   }
    \label{fig:framework}
\end{figure*}
\vspace{-0.5em}

\textbf{Mixture of Experts (MoE).} Recent advances in LLM scaling have highlighted the significant potential of  MoE
~\cite{jacobs1991adaptive,shazeer2017outrageously,lepikhin2020gshard,fedus2022switch,du2022glam,lewis2021base,ainslie2023colt5}. 
Recent models such as Mixtral~\cite{jiang2024mixtral}, DeepSeek-V3~\cite{liu2024deepseek} and Qwen-MoE~\cite{qwen3technicalreport} illustrate that MoE can achieve, and sometimes surpass, the performance of dense models of comparable size while activating only a subset of parameters. This makes MoE a promising direction for enhancing the capacity and efficiency of LLMs. Despite the sparsity-induced FLOPs advantage, MoE models still suffer from massive overall parameter counts and substantial activation memory requirements~\cite{rajbhandari2022deepspeed,chowdhery2023palm}, making edge deployment challenging.

\textbf{Post-training Pruning.}
As LLM parameters grow to billions or more, retraining-based pruning~\cite{guan2021cocopie,zhang2022advancing,yang2023pruning,zhan2021achieving,zhan2024exploring,ma2023llm,zhang2023lottery,wu2022compiler,rtseg,li2022pruning,zhan-etal-2024-rethinking-token} becomes prohibitively expensive. PTP has emerged as a mainstream alternative ~\cite{zhao-etal-2024-pruning,shen2024search,shen2024numerical,shen2025sparse} %
to prune pretrained models with lightweight compensation to reduce accuracy loss, instead of expensive retraining.
Inspired by post-training quantization, some methods~\cite{srinivas2015data,kim2020neuron,yvinec2021red,yvinec2022red++} perform layer-wise calibration to minimize output deviation, with SparseGPT~\cite{frantar2023sparsegpt} extending this to large-scale Transformers. %
Recent works~\cite{ashkboos2024slicegpt,ling2024slimgpt} explore structured PTP to achieve significant acceleration while preserving accuracy performance. 
Moe-pruner~\cite{xie2024moe} prunes MoE LLMs by leveraging routing information to remove underutilized experts, reducing model size while maintaining efficiency and quality. 
Considering the compatibility requirements of subsequent quantization, deployment, and platform adaptation, we ultimately adopt a more structured and widely compatible expert pruning strategy.

\textbf{Post-training Quantization.}
PTQ performs offline quantization directly on pretrained models using only a small amount of calibration data~\cite{liu2021post,fang2020post}, making it particularly practical for large-scale models~\cite{wei2022outlier,yao2022zeroquant,ashkboos2024quarot,liu2024spinquant,shen2025quartdepth}. 
For LLMs, various methods \cite{frantar2022gptq,lin2024awq,ma2024affinequant} are proposed to minimize the quantization errors through outlier selection or learnable scaling, leading to superior performance under 4-bit quantization.
To overcome the accuracy bottleneck at 3-bit or 2-bit, 
QuIP\#~\cite{tseng2024quip}, 
and AQLM~\cite{egiazarian2024extreme} introduce codebook learning, matrix factorization, and lightweight finetuning to mitigate quantization loss at minimal training cost. 
In practice, bitsandbytes~\cite{dettmers2023qlora,dettmers2022gpt3,dettmers20218} provides efficient low-precision kernels for 8-bit, 4-bit, and even 2-bit weight quantization. Llama.cpp~\cite{llama_cpp} explores sub-2-bit quantization for large models using calibration, and lookup-table storage, maintaining performance at extremely low bit-widths, and remains one of the few viable options due to its multi-platform support.

\section{Method}

\textbf{Framework Overview.}
Our target is to compress DeepSeek-V3 with 671B parameters and deploy it on edge platforms with 128GB memory budget. One single strategy can hardly satisfy the requirement. Specifically, extreme low-bit quantization with an average 1.56  bit width  cannot fit the inference memory within the practical limits. Meanwhile, aggressive pruning needs to remove substantial weights (above 90\% sparsity ratio) with significant performance degradation. Thus, to achieve massive memory reduction and retain accuracy and performance, we propose a collaborative compression framework by combining Performance-Aware Expert Pruning~\ref{sec:prune}, Hardware-Aware Activation Adjustment~\ref{sec:hardware} and Mixed-Precision Quantization~\ref{sec:quantization}, as shown in Fig~\ref{fig:framework}.
Our framework has the following steps:
\begin{itemize}
 \item Prune experts by estimating the importance of experts and removing low-contribution experts. 
 \item Adjust  expert activation according to pruning configurations.
 \item Quantize the remaining MoE model to the base low precision.
 \item Estimate the quantization sensitivity of each tensor by temporarily upgrading it to a higher precision.
 \item Select high-sensitive tensors with more significant improvements and upgrade their precisions for better performance.
\end{itemize}

\subsection{Performance-Aware Expert Pruning}
\label{sec:prune}

\begin{figure}[t] 
    \centering
    \includegraphics[width=\linewidth]{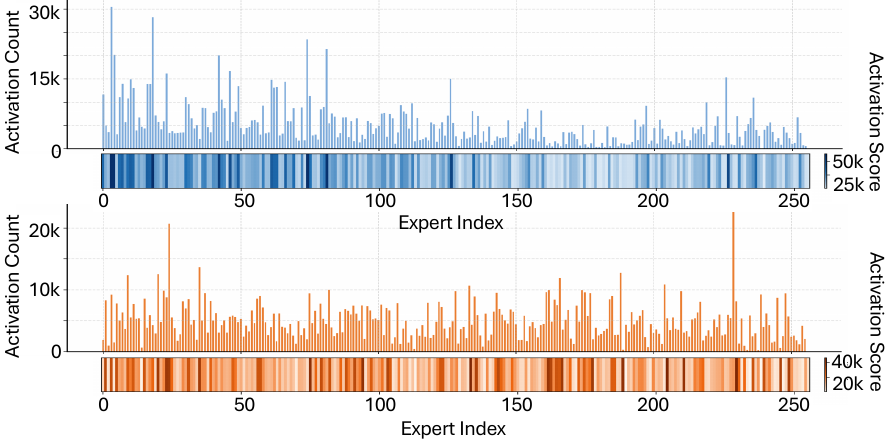}
    \caption{
    Visualization of Expert Activation Count and Cumulative Activation Score in the 1st and last MoE Layer of DeepSeek-V3. Results on the calibration dataset reveal a pronounced imbalance in both activation frequency and contribution across experts, thereby justifying the feasibility of our expert pruning strategy.
    }
    \label{fig:activation_count}
\end{figure}

For DeepSeek-V3~\cite{liu2024deepseek}  with 671 billion parameters and 1.3TB storage, an aggressive quantization with 1.56 bits per weight still leads to 125GB storage cost without accounting for the overhead of intermediate activations, key-value (KV) cache, and communications during inference. To meet the 128GB memory requirement, we first perform expert pruning to reduce the overall model size before quantization.
Specifically, we adopt an expert-level pruning strategy that removes low-contribution experts while preserving the dense structure of activated parameters.
Formally, the importance of each expert $e$ in an MoE layer is measured by a score:
{\small
\[
I_e = \alpha \cdot f_e + (1-\alpha) \cdot \overline{s}_e,
\]}%
where $f_e$ denotes the activation frequency of $e$ over a calibration dataset, $\overline{s}_e$ is the average routing score assigned to $e$ when selected, and $\alpha \in [0,1]$ balances the two terms. The activation frequency and scores are illustrated in Fig.~\ref{fig:activation_count}, demonstrating the activation imbalance and justifying the expert pruning.
Experts are ranked by $I_e$, and the top $\lceil rE \rceil$ experts (for a given retention ratio $r$) are kept.

After pruning, the total parameter count is approximately $P_{\text{pruned}} \approx r \cdot P_{\text{total}}$.
A smaller $r$ yields larger memory savings at the risk of more significant accuracy degradation.
Expert pruning maintains the original model architecture, without additional sparse scheduling overheads that may harm hardware efficiency across CPUs and GPUs. The pruned model can be quantized readily.

\subsection{Hardware-Aware Activation Adjustment}
\label{sec:hardware}

In practice, we find one specific issue that after expert pruning, retaining the original number of activated experts per layer can misroute tokens. Specifically, 
if the number of per-layer activation experts is kept the same as the original value $k_{\text{orig}}$, some tokens which previously are sent to pruned experts will be reassigned to suboptimal ones. This mismatch can accumulate through deep MoE stacks, leading to more severe performance degradation.  To mitigate this, the number of activated experts per-layer is scaled proportionally to the remaining experts. In detail,  we set the per-layer activated experts number to $k_{\text{pruned}} = \lceil rk_{\text{orig}}\rceil$ where $r\in[0,1)$ denotes the retention ratio after pruning and $\lceil\cdot\rceil$ denotes the ceiling to integers. 

Furthermore, on different edge devices with various computation and memory constraints, we can update the number of activated experts to satisfy the limitations. 
We denote $L$ as the number of MoE layers, $k_i$ as the number of activated experts at the $i^{th}$ layer, and $P_i$ as its parameter count of a single expert. 
The computation of active experts per token in the forward pass can be approximated by
{\small
\[
\text{FLOPs}_{\text{active}} \approx 2\sum_{i=1}^{L} k_i\,P_i.
\]
}%
Thus, by changing the value of $k_i$, we can adjust the computations and memory cost.
In practice, we enforce $k_i\le k_{\text{pruned}}$ for all $i$, and tune $k_i$ to respect device-specific compute and memory budgets.
To realize this policy during implementation, two routing/activation hyperparameters are adjusted: (1) \emph{num\_experts\_per\_tok} is set to $k_{\text{new}}$, aligning the per-token activation with the pruned expert pool; and (2) \emph{topk\_group} (grouped top-$k$ routing) is modified to balance routing diversity and locality. Concretely, topk\_group partitions the $E_{\text{pruned}}$ experts into groups and applies a local top-$k$ within each group. This reduces contention and improves cache and communication locality. under tight hardware constraints, while num\_experts\_per\_tok controls the total $k_i$ selected per layer and per token.

With these adjustments, the activation pattern better matches the reduced expert set and hardware limits: (i) Routing-induced mismatches are suppressed. (ii) Peak compute $\text{FLOPs}_{\text{active}}$ and peak activation memory are lowered to device-acceptable ranges. (iii) Core model quality is preserved despite expert pruning.

\subsection{Mixed-Precision Quantization}
\label{sec:quantization}

To satisfy the 128GB memory constraint with an aggressive compression ratio, an extremely low-bit quantization strategy is applied on the pruned model. Since mainstream PTQ methods suffer from significant performance degradation under 3~bits (or below) quantization, we adopt llama.cpp~\cite{llama_cpp}  with extremely low bit-width quantization options for our efficient quantization implementations across multiple platforms including CPUs, GPUs, and edge devices. 

We first quantize the entire model to a base low precision.
To address the performance degradation under extremely low bit-width quantization, we design a mixed precision quantization framework that keeps most weights low bit-width while allowing certain key weights to maintain higher precision. 
To achieve this, a tensor-level sensitivity analysis is proposed to identify the tensor importance. Then more important tensors are selected to keep high precision under the overall memory constraint. 

\textbf{Tensor-Level Sensitivity Analysis.}
We denote $b(Q)$ as the bit-width of quantization scheme $Q$, and $b(Q,t)$ as the bit-width of a specific tensor $t$ under quantization scheme $Q$. The entire model  is first quantized to a base low precision 
$Q_{\text{low}}(\text{IQ1\_M})$,
with a corresponding base perplexity 
$\text{PPL}(Q_{\text{low}})$.
For a tensor $t$ with parameter count $|t|$, we temporarily upgrade it to a higher precision $Q_{\text{high}}\in \{Q8,Q4\}$ while keeping all other tensors at $Q_{\text{low}}$, yielding a new perplexity 
$\text{PPL}(Q_{\text{high}},t)$. $Q8$ denotes quantizing to 8 bits and $Q4$ denotes 4 bit quantization.
The sensitivity of $t$ is defined as
{\small
\[
\text{Sens}(t)=\frac{\text{PPL}(Q_{\text{low}})-\text{PPL}(Q_{\text{high}},t)}{\rho_t},
\]
}%
where $\rho_t=|t|/|\mathrm{layer}(t)|$ denotes the proportion of parameters in $t$ relative to its Transformer layer. 
A higher $\text{Sens}(t)$ indicates greater accuracy benefit per unit parameter when increasing $t$'s bit-width.

\textbf{Hardware-Constrained Budget Formulation.}
Let $P_{\text{pruned}}$ be the total parameter count after expert pruning (Section~\ref{sec:prune}) with a memory cost of $S_{\text{low}}=P_{\text{pruned}}\cdot b(Q_{\text{low}})/{(8\cdot 10^9)} \quad \text{(GB)}$, where dividing $8\cdot 10^9$ converts the scale to Gigabyte. Given the memory constraint $M_{\text{limit}}$ (GB), the base memory budget is $B_{\text{base}} = M_{\text{limit}} - S_{\text{low}}$, which can be used to improve certain weight precision.

Additional budget $B_{\delta}$ can be released by further lowering the precision of low-sensitivity tensors: downgrading a tensor $t$ from bit-width $b(Q_{\text{low}},t)$ to $b(Q_{\text{low}'},t)$ frees $B_{\delta}$ (GB) memory as below,
{\small
\[B_{\delta}=|t|\cdot \big(b(Q_{\text{low}},t)-b(Q_{\text{low}'},t)\big)/({8\cdot 10^9}).\] 
}%
The total upgrade budget is then $B_{\text{total}}=B_{\text{base}}+B_{\delta}$.

\textbf{Dynamic Allocation with Back-off.}
Next based on the sensitivity analysis of all tensors and the available memory budget, we select certain more quantization-sensitive (more important) tensors and upgrade their precisions to maintain model performance. Specifically, 
tensors are sorted in descending order of $\text{Sens}(t)$ and upgraded from $Q_{\text{low}}$ toward higher precisions. 
Specifically, tensors with fewer than 10 million parameters are  upgraded  to Q8 precision. For medium-sized tensors with 10 million to 50 million parameters, their precision are increased to Q6. 
Routed experts with billions of parameters are not permitted for precision upgradation and can only be downgraded during back-off, due to their enormous memory footprint.

Let $T_{\uparrow}$ be the set of upgraded tensors. The upgrade plan must satisfy the constraint
{\small
\[
\sum_{t\in T_{\uparrow}} |t|\cdot \big({b(Q_{\text{new}},t)-b(Q_{\text{low}},t)}\big)
/({8\cdot 10^9}) \ \le\ B_{\text{total}}, %
\]
}%
where $b(Q_{\text{new}},t)$ is the upgraded bit-width of $t$. 
If the budget constraint is violated during upgrading, we adopt a deterministic back-off  with the following priority rules:  
(1) \textbf{Local downgrade}: We first downgrade the current tensor by only one level of precision (e.g., $Q8\!\rightarrow\!Q6$). If reducing its precision by two levels is still insufficient to meet the budget, proceed to the next step.  
(2) \textbf{Global large-tensor downgrade}: Among all previously upgraded tensors, we select the one with the most parameters and downgrade it by one level. For highly sensitive tensors that have been upgraded to $Q8$, at most three downgrades are allowed (i.e., $Q8\!\rightarrow\!Q4$ ) to avoid excessive loss of accuracy.  
(3) \textbf{Routed expert downgrade}: if the violation persists, we downgrade routed experts from $\text{IQ1\_M}$ to the lowest level $\text{IQ1\_S}$. 
The downgraded layer is selected following  a binary-partition order (e.g., in a 58-layer MoE, the sequence is layer 29, 15, 44, 8, and so on), and ffn\_gate \& ffn\_up tensors are downgraded first before other.

\begin{figure}[t] 
    \centering
    \includegraphics[width=\linewidth]{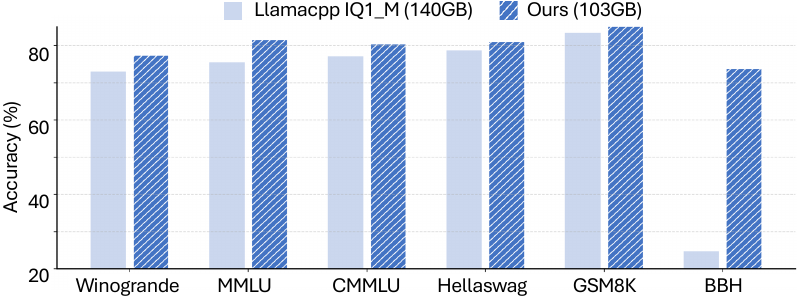}
    \caption{DeepSeek-V3 benchmark comparision of llamacpp unified IQ1\_M quant (140G) and ours smallest compression 103GB model.}
    \label{fig:barchart}
\end{figure}

\section{Experiments}

\subsection{Experiments Settings}

\textbf{Models, Datasets and Evaluation. }
We conduct experiments on the DeepSeek-V3~\cite{liu2024deepseek} MoE LLM.
For pruning and quantization calibration, we employ three representative datasets. We use subset of
\textit{MMLU}~\cite{hendrycks2020measuring}, %
and two open source calibration dataset \cite{bartowski_calibration_data, bartowski_gguf_calibration}.
Perplexity is measured on \textit{WikiText-2-raw}~\cite{merity2016pointer} using the default llama.cpp setup. For downstream evaluation after pruning and quantization, we adopt eight benchmarks: \textit{MMLU}~\cite{hendrycks2020measuring} and \textit{CMMLU}~\cite{li2023cmmlu} for multi-task reasoning, \textit{GSM8K}~\cite{cobbe2021training} for math problem solving, \textit{BBH}~\cite{suzgun2022challenging} for broad reasoning, \textit{WinoGrande}~\cite{sakaguchi2021winogrande} and \textit{HellaSwag}~\cite{zellers2019hellaswag} for commonsense and sentence completion, \textit{AIME24}~\cite{aime_2024} and \textit{GPQA}~\cite{rein2023gpqa} for advanced reasoning.

\begin{table}[t]
  \centering
  \caption{Benchmark results of full-seized DeepSeek-V3(671B) between ours mixed precision optimized version and 2 baseline quant.}
  \label{tab:benchmarks}
  \renewcommand{\arraystretch}{1.2} %
  \setlength{\tabcolsep}{2 pt}      %
  \resizebox{\columnwidth}{!}{%
  \begin{tabular}{l | c | ccccc}
    \toprule
    Model/Benchmark & Size & Winogrande & MMLU & CMMLU & HellaSwag & GSM8K \\
    \midrule
   IQ1\_M (uniform)   & 140GB 
    & 73    & 75.45 & 77.06 & 78.7  & 83.4  \\
 Q2\_K\_L (uniform)
    & 230GB 
    & \textbf{77.74} & \textbf{85.57} & 82.57 & \textbf{86.46} & 93.4  \\
    \rowcolor{RowColor}
    Ours (Mixed)                & \textbf{130GB} 
    & 77.63 & 84.55 & \textbf{86.08} & 84.58 & \textbf{93.56} \\
    \bottomrule
  \end{tabular}%
  }
\end{table}

\textbf{Baseline and Testing Methods. }
We apply the proposed framework—comprising expert pruning, MoE-specific quantization, and activation optimization—to  DeepSeek-V3~\cite{liu2024deepseek}, and compare with multiple baselines including (1) the original model quantized with default Q2\_K\_L method, and (2) the original model quantized with default IQ1\_M method. %
All experiments begin with BF16-precision evaluation and expert pruning on an 8$\times$NVIDIA~H200 server. %
For low-bit quantization, we leverage the llama.cpp importance-matrix method tailored for specific ultra-low-bit formats, performing importance-group calibration on an 8$\times$NVIDIA~H100~NVL server. We then apply the mixed-precision allocation strategy to the pruned and activation-optimized model.
To ensure feasibility on the edge, %
we deploy compressed models on an AMD~Ryzen~AI~Max+ ``Strix Halo'' laptop, measuring both latency and throughput. %

\textbf{Implementation Details.}In our implementation of Expert Reduction, the weighting factor $\alpha$ is set to 0.5 to balance the influence of activation count and activation score. The pruning ratio \emph{r} was set to $0.75$, which we found to provide the best trade-off between maximum compression and reasonable performance. After pruning, the model retains \emph{508B} parameters. During activation adjustment, we set \emph{num\_experts\_per\_tok} to 6 (out of 192) to reduce the computational overhead while partially compensating for the performance degradation caused by pruning, resulting in \emph{32B} activated parameters during inference.  For quantization, all \emph{norm}-related weights are kept in \emph{F32}. \emph{k\_b, v\_b} in all attention blocks, \emph{token\_embd} and the final \emph{output} are the only tensors quantized to $Q8$.

\begin{table}[t]
  \centering
  \footnotesize
  \setlength{\tabcolsep}{3pt}
  \renewcommand{\arraystretch}{1.2}
  \caption{Benchmark result of Qwen3 MoE(235B) at 8bit and our DeepSeek-R1(671B) quant at similart size across resoning datasets.}
  \label{tab:r1_benchmark}
  \resizebox{\columnwidth}{!}{%
  \begin{tabular}{l|l|l|c|c|c}
    \hline
    Model & Params & Size & AIME24 & GPQA\_Diamond & GSM8K \\
    \hline
    Qwen3 (8bit) & 235B/22B & 233GB & 86.67 & 68.18 & 84.99 \\
    \rowcolor{RowColor}
    R1 (Ours)    & 671B/37B & \textbf{210GB} & \textbf{86.67} & \textbf{74.24} & \textbf{96.51} \\
    \hline
  \end{tabular}
  }
\end{table}

\subsection{Results}

\textbf{Effective Compression to Meet Requirements.}
With our proposed collaborative compression framework, we reduce the memory storage of DeepSeek-V3  from 1.3TB in BF16)
to an unprecedented 103GB, which can be deployed on AMD’s latest RyzenAIMax+ ``Strix Halo'' laptop with a total of 128GB memory. Such an extreme compression ratio was previously considered infeasible,  as evidenced by the entirely non-functional smallest available DeepSeek-V3 under the IQ1\_S quantization in llama.cpp with 130GB storage.
To demonstrated the model performance after compression, we compare with the uniform quantization under IQ1\_M  for DeepSeek-V3 in llama.cpp. As shown in Fig~\ref{fig:barchart}, our method leads to not only smaller storage (our 103GB v.s. 140GB from uniform quantization), but also higher accuracy across various benchmarks spanning diverse capabilities, demonstrating the effectiveness of our method for practical deployment. Notably, on BBH for broad reasoning evaluation, the 140GB model suffers from significant accuracy degradation. However, our 103GB model can still maintain higher accuracy on BBH with strong reasoning capabilities. 

\textbf{Higher Accuracy with Smaller Sizes.}
We conduct an ablation study to further validate the effectiveness of our mixed-precision quantization. Specifically, we compare our mixed-precision quantization with uniform low-bit quantization on the full-size DeepSeek-V3 model. As shown in Table~\ref{tab:benchmarks}, under an average bit-width below 2, compared with the uniform IQ1\_M quantization, our mixed method achieves superior accuracy with a smaller storage size. The accuracy is even comparable with that of the 230GB Q2-level (2.78bpw with Q2\_K\_L) quantization, which costs about 1.8$\times$ storage.
Furthermore, we evaluate our mixed-precision quantization for DeepSeek-R1, and compare with the 8-bit Qwen3 MoE under comparable storage size.  As presented in Table~\ref{tab:r1_benchmark}, under about 220GB size, our quantized R1 model exhibits a clear advantage across various benchmarks with a slightly smaller storage cost, compared with the 8-bit Qwen3 MoE. In summary, these experiments comprehensively demonstrate the effectiveness and superiority of our proposed mixed-precision quantization scheme.

\textbf{Practical Real-World Deployment.}
Finally, we validate the effectiveness of our compression framework through real-world deployment on an AI PC platform. Specifically, we select AMD’s latest Ryzen AI Max+ ‘Strix Halo’ laptop, equipped with 128GB of total memory, including 64GB dedicated VRAM, 32GB shared VRAM, and 32GB RAM. After model deployment and system-level optimization, we successfully deployed and ran our compressed 103GB DeepSeek-V3 MoE model on this lightweight laptop, achieving a generation speed of over 5~tokens/s. This real-world experiment provides strong evidence for our feasibility and effectiveness. %

\section{Conclusion}
We proposed a collaborative compression framework that integrates expert pruning, mixed-precision quantization, and activation optimization. %
Our approach enables, to the best of our knowledge, the first successful deployment of DeepSeek-V3-scale MoE models under a 128~GB memory budget. Extensive experiments further demonstrate that our method consistently outperforms uniform low-bit quantization techniques, achieving better accuracy while lowering memory usage.

\bibliographystyle{IEEEbib}
\bibliography{refs}

\begin{thebibliography}{10}

\bibitem{jacobs1991adaptive}
Robert~A Jacobs, Michael~I Jordan, et~al.,
\newblock ``Adaptive mixtures of local experts,''
\newblock {\em Neural Comput.}, 1991.

\bibitem{shazeer2017outrageously}
Noam Shazeer et~al.,
\newblock ``Outrageously large neural networks: The sparsely-gated mixture-of-experts layer,''
\newblock {\em arXiv preprint arXiv:1701.06538}, 2017.

\bibitem{lepikhin2020gshard}
Dmitry Lepikhin, HyoukJoong Lee, et~al.,
\newblock ``Gshard: Scaling giant models with conditional computation and automatic sharding,''
\newblock {\em arXiv preprint arXiv:2006.16668}, 2020.

\bibitem{fedus2022switch}
William Fedus et~al.,
\newblock ``Switch transformers: Scaling to trillion parameter models with simple and efficient sparsity,''
\newblock {\em JMLR}, 2022.

\bibitem{jiang2024mixtral}
Albert~Q Jiang, Alexandre Sablayrolles, et~al.,
\newblock ``Mixtral of experts,''
\newblock {\em arXiv preprint arXiv:2401.04088}, 2024.

\bibitem{liu2024deepseek}
Aixin Liu, Bei Feng, et~al.,
\newblock ``Deepseek-v3 technical report,''
\newblock {\em arXiv preprint arXiv:2412.19437}, 2024.

\bibitem{team2025kimi}
Kimi Team, Yifan Bai, et~al.,
\newblock ``Kimi k2: Open agentic intelligence,''
\newblock {\em arXiv preprint arXiv:2507.20534}, 2025.

\bibitem{qwen3technicalreport}
Qwen Team et~al.,
\newblock ``Qwen3 technical report,''
\newblock {\em arXiv preprint arXiv:2505.09388}, 2025.

\bibitem{rajbhandari2022deepspeed}
Samyam Rajbhandari et~al.,
\newblock ``Deepspeed-moe: Advancing mixture-of-experts inference and training to power next-generation ai scale,''
\newblock in {\em ICML}, 2022.

\bibitem{he2022fastermoe}
Jiaao He, Jidong Zhai, et~al.,
\newblock ``Fastermoe: Modeling and optimizing training of large-scale dynamic pre-trained models,''
\newblock in {\em PPoPP}, 2022.

\bibitem{hwang2023tutel}
Changho Hwang, Wei Cui, et~al.,
\newblock ``Tutel: Adaptive mixture-of-experts at scale,''
\newblock {\em MLSys}, 2023.

\bibitem{frantar2022gptq}
Elias Frantar et~al.,
\newblock ``Gptq: Accurate post-training quantization for generative pre-trained transformers,''
\newblock {\em arXiv preprint arXiv:2210.17323}, 2022.

\bibitem{lin2024awq}
Ji~Lin, Jiaming Tang, et~al.,
\newblock ``Awq: Activation-aware weight quantization for on-device llm compression and acceleration,''
\newblock {\em MLSys}, 2024.

\bibitem{lee2024owq}
Changhun Lee et~al.,
\newblock ``Owq: Outlier-aware weight quantization for efficient fine-tuning and inference of large language models,''
\newblock in {\em AAAI}, 2024.

\bibitem{ma2024affinequant}
Yuexiao Ma et~al.,
\newblock ``Affinequant: Affine transformation quantization for large language models,''
\newblock {\em arXiv preprint arXiv:2403.12544}, 2024.

\bibitem{llama_cpp}
Georgi Gerganov et~al.,
\newblock ``ggml-org/llama.cpp,'' \url{https://github.com/ggerganov/llama.cpp}, 2023.

\bibitem{xiao2023smoothquant}
Guangxuan Xiao, Ji~Lin, et~al.,
\newblock ``Smoothquant: Accurate and efficient post-training quantization for large language models,''
\newblock in {\em ICML}, 2023.

\bibitem{wei2022qdrop}
Xiuying Wei et~al.,
\newblock ``Qdrop: Randomly dropping quantization for extremely low-bit post-training quantization,''
\newblock {\em arXiv preprint arXiv:2203.05740}, 2022.

\bibitem{dao2022flashattention}
Tri Dao, Dan Fu, et~al.,
\newblock ``Flashattention: Fast and memory-efficient exact attention with io-awareness,''
\newblock in {\em NeurIPS}, 2022.

\bibitem{dao2023flashattention}
Tri Dao,
\newblock ``Flashattention-2: Faster attention with better parallelism and work partitioning,''
\newblock {\em arXiv preprint arXiv:2307.08691}, 2023.

\bibitem{ren2021zero}
Jie Ren, Samyam Rajbhandari, Reza~Yazdani Aminabadi, Olatunji Ruwase, Shuangyan Yang, Minjia Zhang, Dong Li, and Yuxiong He,
\newblock ``$\{$Zero-offload$\}$: Democratizing $\{$billion-scale$\}$ model training,''
\newblock in {\em 2021 USENIX Annual Technical Conference (USENIX ATC 21)}, 2021, pp. 551--564.

\bibitem{shen2024lazydit}
Xuan Shen, Zhao Song, Yufa Zhou, et~al.,
\newblock ``Lazydit: Lazy learning for the acceleration of diffusion transformers,''
\newblock in {\em AAAI}, 2025.

\bibitem{zhan2024fast}
Zheng Zhan, Yushu Wu, Yifan Gong, et~al.,
\newblock ``Fast and memory-efficient video diffusion using streamlined inference,''
\newblock in {\em NeurIPS}, 2024.

\bibitem{chen2016training}
Tianqi Chen, Bing Xu, et~al.,
\newblock ``Training deep nets with sublinear memory cost,''
\newblock {\em arXiv preprint arXiv:1604.06174}, 2016.

\bibitem{du2022glam}
Nan Du, Yanping Huang, et~al.,
\newblock ``Glam: Efficient scaling of language models with mixture-of-experts,''
\newblock in {\em ICML}, 2022.

\bibitem{lewis2021base}
Mike Lewis, Shruti Bhosale, et~al.,
\newblock ``Base layers: Simplifying training of large, sparse models,''
\newblock in {\em ICML}, 2021.

\bibitem{ainslie2023colt5}
Joshua Ainslie, Tao Lei, et~al.,
\newblock ``Colt5: Faster long-range transformers with conditional computation,''
\newblock {\em arXiv preprint arXiv:2303.09752}, 2023.

\bibitem{chowdhery2023palm}
Aakanksha Chowdhery, Sharan Narang, et~al.,
\newblock ``Palm: Scaling language modeling with pathways,''
\newblock {\em JMLR}, 2023.

\bibitem{guan2021cocopie}
Hui Guan, Shaoshan Liu, et~al.,
\newblock ``Cocopie: Enabling real-time ai on off-the-shelf mobile devices via compression-compilation co-design,''
\newblock {\em Communications of the ACM}, vol. 64, no. 6, pp. 62--68, 2021.

\bibitem{zhang2022advancing}
Yihua Zhang, Yuguang Yao, Parikshit Ram, et~al.,
\newblock ``Advancing model pruning via bi-level optimization,''
\newblock {\em NeurIPS}, 2022.

\bibitem{yang2023pruning}
Changdi Yang, Pu~Zhao, Yanyu Li, et~al.,
\newblock ``Pruning parameterization with bi-level optimization for efficient semantic segmentation on the edge,''
\newblock in {\em CVPR}, 2023.

\bibitem{zhan2021achieving}
Zheng Zhan, Yifan Gong, Pu~Zhao, Geng Yuan, et~al.,
\newblock ``Achieving on-mobile real-time super-resolution with neural architecture and pruning search,''
\newblock in {\em ICCV}, 2021, pp. 4821--4831.

\bibitem{zhan2024exploring}
Zheng Zhan, Zhenglun Kong, Yifan Gong, et~al.,
\newblock ``Exploring token pruning in vision state space models,''
\newblock in {\em NeurIPS}, 2024.

\bibitem{ma2023llm}
Xinyin Ma, Gongfan Fang, et~al.,
\newblock ``Llm-pruner: On the structural pruning of large language models,''
\newblock {\em NeurIPS}, 2023.

\bibitem{zhang2023lottery}
Yuxin Zhang, Mingbao Lin, et~al.,
\newblock ``Lottery jackpots exist in pre-trained models,''
\newblock {\em IEEE TPAMI}, 2023.

\bibitem{wu2022compiler}
Yushu Wu, Yifan Gong, Pu~Zhao, et~al.,
\newblock ``Compiler-aware neural architecture search for on-mobile real-time super-resolution,''
\newblock in {\em ECCV}. Springer, 2022, pp. 92--111.

\bibitem{rtseg}
Yanyu Li, Changdi Yang, Pu~Zhao, et~al.,
\newblock ``Towards real-time segmentation on the edge,''
\newblock 2023, AAAI'23/IAAI'23/EAAI'23.

\bibitem{li2022pruning}
Yanyu Li, Pu~Zhao, Geng Yuan, Xue Lin, Yanzhi Wang, and Xin Chen,
\newblock ``Pruning-as-search: Efficient neural architecture search via channel pruning and structural reparameterization,''
\newblock {\em arXiv preprint arXiv:2206.01198}, 2022.

\bibitem{zhan-etal-2024-rethinking-token}
Zheng Zhan, Yushu Wu, Zhenglun Kong, et~al.,
\newblock ``Rethinking token reduction for state space models,''
\newblock in {\em EMNLP}, Miami, Florida, USA, nov 2024, pp. 1686--1697, ACL.

\bibitem{zhao-etal-2024-pruning}
Pu~Zhao, Fei Sun, Xuan Shen, et~al.,
\newblock ``Pruning foundation models for high accuracy without retraining,''
\newblock in {\em Findings of EMNLP 2024}. 2024, ACL.

\bibitem{shen2024search}
Xuan Shen, Pu~Zhao, et~al.,
\newblock ``Search for efficient large language models,''
\newblock in {\em NeurIPS}, 2024.

\bibitem{shen2024numerical}
Xuan Shen, Zhao Song, Yufa Zhou, et~al.,
\newblock ``Numerical pruning for efficient autoregressive models,''
\newblock in {\em AAAI}, 2025.

\bibitem{shen2025sparse}
Xuan Shen, Hangyu Zheng, Yifan Gong, et~al.,
\newblock ``Sparse learning for state space models on mobile,''
\newblock in {\em ICLR}, 2025.

\bibitem{srinivas2015data}
Suraj Srinivas, R~Venkatesh Babu, et~al.,
\newblock ``Data-free parameter pruning for deep neural networks,''
\newblock {\em arXiv preprint arXiv:1507.06149}, 2015.

\bibitem{kim2020neuron}
Woojeong Kim, Suhyun Kim, et~al.,
\newblock ``Neuron merging: Compensating for pruned neurons,''
\newblock {\em NeurIPS}, 2020.

\bibitem{yvinec2021red}
Edouard Yvinec, Arnaud Dapogny, et~al.,
\newblock ``Red: Looking for redundancies for data-free structured compression of deep neural networks,''
\newblock {\em NeurIPS}, 2021.

\bibitem{yvinec2022red++}
Edouard Yvinec, Arnaud Dapogny, et~al.,
\newblock ``Red++: Data-free pruning of deep neural networks via input splitting and output merging,''
\newblock {\em IEEE TPAMI}, 2022.

\bibitem{frantar2023sparsegpt}
Elias Frantar and Dan Alistarh,
\newblock ``Sparsegpt: Massive language models can be accurately pruned in one-shot,''
\newblock in {\em ICML}, 2023.

\bibitem{ashkboos2024slicegpt}
Saleh Ashkboos et~al.,
\newblock ``Slicegpt: Compress large language models by deleting rows and columns,''
\newblock {\em arXiv preprint arXiv:2401.15024}, 2024.

\bibitem{ling2024slimgpt}
Gui Ling et~al.,
\newblock ``Slimgpt: Layer-wise structured pruning for large language models,''
\newblock {\em NeurIPS}, 2024.

\bibitem{xie2024moe}
Yanyue Xie et~al.,
\newblock ``Moe-pruner: Pruning mixture-of-experts large language model using the hints from its router,''
\newblock {\em arXiv preprint arXiv:2410.12013}, 2024.

\bibitem{liu2021post}
Zhenhua Liu et~al.,
\newblock ``Post-training quantization for vision transformer,''
\newblock {\em NeurIPS}, 2021.

\bibitem{fang2020post}
Jun Fang et~al.,
\newblock ``Post-training piecewise linear quantization for deep neural networks,''
\newblock in {\em ECCV}, 2020.

\bibitem{wei2022outlier}
Xiuying Wei et~al.,
\newblock ``Outlier suppression: Pushing the limit of low-bit transformer language models,''
\newblock {\em NeurIPS}, 2022.

\bibitem{yao2022zeroquant}
Zhewei Yao, Reza Yazdani~Aminabadi, et~al.,
\newblock ``Zeroquant: Efficient and affordable post-training quantization for large-scale transformers,''
\newblock {\em NeurIPS}, vol. 35, pp. 27168--27183, 2022.

\bibitem{ashkboos2024quarot}
Saleh Ashkboos, Amirkeivan Mohtashami, et~al.,
\newblock ``Quarot: Outlier-free 4-bit inference in rotated llms,''
\newblock {\em NeurIPS}, vol. 37, pp. 100213--100240, 2024.

\bibitem{liu2024spinquant}
Zechun Liu et~al.,
\newblock ``Spinquant: Llm quantization with learned rotations,''
\newblock {\em arXiv preprint arXiv:2405.16406}, 2024.

\bibitem{shen2025quartdepth}
Xuan Shen, Weize Ma, Jing Liu, et~al.,
\newblock ``Quartdepth: Post-training quantization for real-time depth estimation on the edge,''
\newblock in {\em CVPR}, 2025.

\bibitem{tseng2024quip}
Albert Tseng et~al.,
\newblock ``Quip\#: Even better llm quantization with hadamard incoherence and lattice codebooks,''
\newblock {\em arXiv preprint arXiv:2402.04396}, 2024.

\bibitem{egiazarian2024extreme}
Vage Egiazarian et~al.,
\newblock ``Extreme compression of large language models via additive quantization,''
\newblock {\em arXiv preprint arXiv:2401.06118}, 2024.

\bibitem{dettmers2023qlora}
Tim Dettmers, Artidoro Pagnoni, et~al.,
\newblock ``Qlora: Efficient finetuning of quantized llms,''
\newblock {\em NeurIPS}, 2023.

\bibitem{dettmers2022gpt3}
Tim Dettmers et~al.,
\newblock ``Gpt3.int8(): 8-bit matrix multiplication for transformers at scale,''
\newblock {\em NeurIPS}, 2022.

\bibitem{dettmers20218}
Tim Dettmers, Mike Lewis, Sam Shleifer, and Luke Zettlemoyer,
\newblock ``8-bit optimizers via block-wise quantization,''
\newblock {\em arXiv preprint arXiv:2110.02861}, 2021.

\bibitem{hendrycks2020measuring}
Dan Hendrycks et~al.,
\newblock ``Measuring massive multitask language understanding,''
\newblock {\em arXiv preprint arXiv:2009.03300}, 2020.

\bibitem{bartowski_calibration_data}
Bartowski et~al.,
\newblock ``Calibration data used for qwen3,'' \url{https://gist.github.com/bartowski1182/f003237f2e8612278a6d01622af1cb6f}, 2024.

\bibitem{bartowski_gguf_calibration}
Bartowski et~al.,
\newblock ``Calibration data v3 (for gguf imatrix files),'' \url{https://gist.github.com/bartowski1182/eb213dccb3571f863da82e99418f81e8}, 2024.

\bibitem{merity2016pointer}
Stephen Merity et~al.,
\newblock ``Pointer sentinel mixture models,''
\newblock {\em arXiv preprint arXiv:1609.07843}, 2016.

\bibitem{li2023cmmlu}
Haonan Li et~al.,
\newblock ``Cmmlu: Measuring massive multitask language understanding in chinese,''
\newblock {\em arXiv preprint arXiv:2306.09212}, 2023.

\bibitem{cobbe2021training}
Karl Cobbe et~al.,
\newblock ``Training verifiers to solve math word problems,''
\newblock {\em arXiv preprint arXiv:2110.14168}, 2021.

\bibitem{suzgun2022challenging}
Mirac Suzgun et~al.,
\newblock ``Challenging big-bench tasks and whether chain-of-thought can solve them,''
\newblock {\em arXiv preprint arXiv:2210.09261}, 2022.

\bibitem{sakaguchi2021winogrande}
Keisuke Sakaguchi et~al.,
\newblock ``Winogrande: An adversarial winograd schema challenge at scale,''
\newblock {\em CACM}, 2021.

\bibitem{zellers2019hellaswag}
Rowan Zellers et~al.,
\newblock ``Hellaswag: Can a machine really finish your sentence?,''
\newblock {\em arXiv preprint arXiv:1905.07830}, 2019.

\bibitem{aime_2024}
Maxwell Jia,
\newblock ``Aime problem set 2024,'' 2024.

\bibitem{rein2023gpqa}
David Rein et~al.,
\newblock ``Gpqa: A graduate-level google-proof q\&a benchmark,'' 2023.

\end{thebibliography}
\balance

\end{document}